\documentclass[10pt,twocolumn,letterpaper]{article}

\usepackage[pagenumbers]{cvpr} %

\usepackage{graphicx}
\usepackage{amsmath}
\usepackage{amssymb}
\usepackage{booktabs}
\usepackage[accsupp]{axessibility}
\usepackage[T1]{fontenc}
\usepackage[latin9]{inputenc}
\usepackage{array}
\usepackage{bbding}
\usepackage{multirow}
\usepackage{balance}

\usepackage{appendix}

\makeatletter

\usepackage[pagebackref,breaklinks,colorlinks]{hyperref}

\usepackage[capitalize]{cleveref}
\crefname{section}{Sec.}{Secs.}
\Crefname{section}{Section}{Sections}
\Crefname{table}{Table}{Tables}
\crefname{table}{Tab.}{Tabs.}

\usepackage{appendix}

\begin{document}

\title{ZegCLIP: Towards Adapting CLIP for Zero-shot Semantic Segmentation}

\author{
    Ziqin Zhou\textsuperscript{\rm 1}\quad\quad
    Yinjie Lei\textsuperscript{\rm 2}\quad\quad
    Bowen Zhang\textsuperscript{\rm 1}\quad\quad
    Lingqiao Liu\textsuperscript{\rm 1}\thanks{Corresponding author}\quad\quad
    Yifan Liu\textsuperscript{\rm 1}
    \and
    {\textsuperscript{\rm 1}The University of Adelaide, Australia}\quad\quad
    {\textsuperscript{\rm 2}Sichuan University, China}
    \and
    {\tt\small \{ziqin.zhou,b.zhang,lingqiao.liu,yifan.liu04\}@adelaide.edu.au,yinjie@scu.edu.cn}
}

\maketitle

\pagestyle{empty} %
\thispagestyle{empty} %

\begin{abstract}
Recently, CLIP has been applied to pixel-level zero-shot learning tasks via a \textbf{two-stage} scheme. The general idea is to first generate class-agnostic region proposals and then feed the cropped proposal regions to CLIP to utilize its image-level zero-shot classification capability. While effective, such a scheme requires two image encoders, one for proposal generation and one for CLIP, leading to a complicated pipeline and high computational cost. In this work, we pursue a simpler-and-efficient \textbf{one-stage} solution that directly extends CLIP's zero-shot prediction capability from image to pixel level. Our investigation starts with a straightforward extension as our baseline that generates semantic masks by comparing the similarity between text and patch embeddings extracted from CLIP. However, such a paradigm could heavily overfit the seen classes and fail to generalize to unseen classes. To handle this issue, we propose three simple-but-effective designs and figure out that they can significantly retain the inherent zero-shot capacity of CLIP and improve pixel-level generalization ability. Incorporating those modifications leads to an efficient zero-shot semantic segmentation system called \textbf{ZegCLIP}. Through extensive experiments on three public benchmarks, ZegCLIP demonstrates superior performance, outperforming the state-of-the-art methods by a large margin under both ``inductive'' and ``transductive'' zero-shot settings. In addition, compared with the two-stage method, our one-stage ZegCLIP achieves a speedup of about \textbf{5 times faster} during inference. We release the code at \url{https://github.com/ZiqinZhou66/ZegCLIP.git}.

\end{abstract}

\vspace{-5mm}
\section{Introduction}\label{sec:intro}

\begin{figure}[t]
\begin{center}
\includegraphics[width=1.0\linewidth]{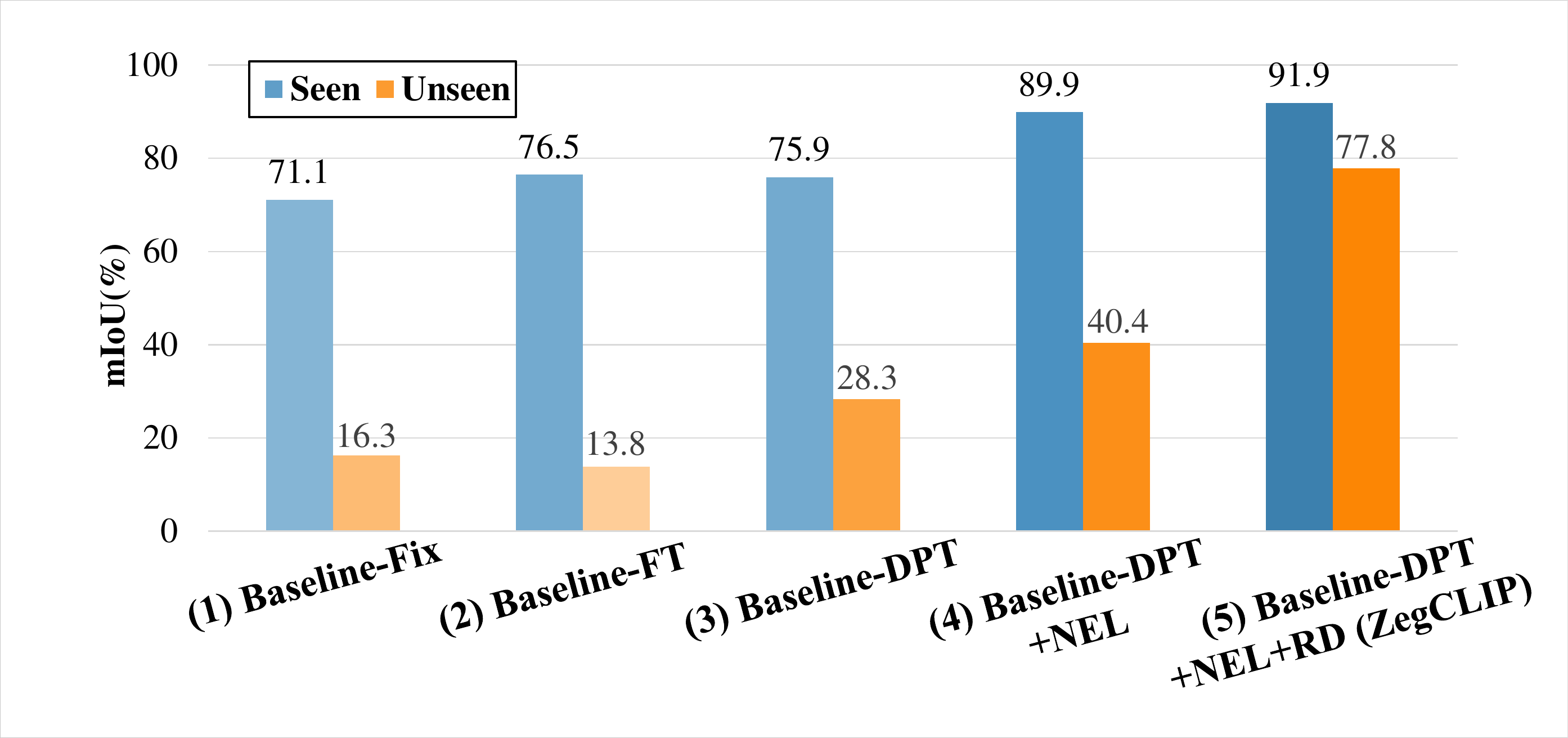}
\end{center}
\vspace{-5mm}
\caption{Quantitative improvements achieved by our proposed designs on VOC dataset. (1)(2) represents our one-stage \textbf{Baseline} model of different versions (Fix or Fine-Tune CLIP image encoder), while (3)-(5) shows the effectiveness of applying our proposed designs, i.e., \textbf{Deep Prompt Tuning (DPT)}, \textbf{Non-mutually Exclusive Loss (NEL)}, \textbf{Relationship Descriptor (RD)}, on baseline model step by step. We highlight that our designs can dramatically increase the segmentation performance on unseen classes.}
\vspace{-5mm}
\label{fig:effectiveness}
\end{figure}

Semantic segmentation is one of the fundamental tasks in the computer vision field, which aims to predict the category of each pixel of an image  \cite{van2021unsupervised,fan2021rethinking,chen2021semi,melas2022deep}. Extensive works have been proposed \cite{liu2019auto,lu2020video,cheng2022masked}, e.g., Fully Convolutional Networks \cite{long2015fully}, U-net \cite{ronneberger2015unet}, DeepLab family \cite{chen2014deeplab1,chen2017deeplab2,chen2017deeplab3} and more recently Vision Transformer based methods \cite{zhu2021unified,gu2022multi,zhang2022topformer}. However, the success of the deep semantic segmentation models heavily relies on the availability of a large amount of annotated training images, which involves a substantial amount of labor. This gives rise to a surging interest in low-supervision-based semantic segmentation approaches, including semi-supervised \cite{chen2021semi}, weakly-supervised \cite{xu2021leveraging}, few-shot \cite{xie2021scale}, and zero-shot semantic segmentation\cite{bucher2019zero,pastore2021closer,xian2019semantic}.

Among them, zero-shot semantic segmentation is particularly challenging and attractive since it is required to directly produce the semantic segmentation results based on the semantic description of a given class. Recently, the pre-trained vision-language model CLIP \cite{radford2021learning} has been adopted into various dense prediction tasks, such as referring segmentation \cite{wang2022cris}, semantic segmentation \cite{pakhomov2021segmentation}, and detection \cite{esmaeilpour2022zero}. It also offers a new paradigm and has made a breakthrough for zero-shot semantic segmentation. Initially built for matching text and images, CLIP has demonstrated a remarkable capability for image-level zero-shot classification.  

However, zsseg \cite{xu2021simple} and Zegformer\cite{ding2022decoupling} follow the common strategy that needs \textbf{two-stage} processing that first generates region proposals and then feeds the cropped regions to CLIP for zero-shot classification.
Such a strategy requires two image encoding processes, one for generating proposals and one for encoding each proposal via CLIP. This design creates additional computational overhead and cannot leverage the knowledge of the CLIP encoder at the proposal generation stage. Besides, MaskCLIP+\cite{zhou2022maskclip} utilizes CLIP to generate pseudo labels of novel classes for self-training but will be invalid if the unseen class names in inference are unknown in the training stage (``inductive'' zero-shot setting).

This paper pursues simplifying the pipeline by directly extending the zero-shot capability of CLIP from image-level to pixel-level.
The basic idea of our method is straightforward: we use a lightweight decoder to match the text prompts against the local embeddings extracted from CLIP, which could be achieved via the self-attention mechanism in a transformer-based structure. We train the vanilla decoder and fix or fine-tune the CLIP image encoder on a dataset containing pixel-level annotations from a limited number of classes, expecting the text-patch matching capability can generalize to unseen classes. Unfortunately, this basic version tends to overfit the training set: while the segmentation results for seen classes generally improve, the model fails to produce reasonable segments on unseen classes. Surprisingly, we discover such an overfitting issue can be dramatically alleviated by incorporating three modified design choices
and report the quantitative improvements in Fig.~\ref{fig:effectiveness}. The following highlights our key discoveries:

\begin{table}[t]
\caption{Differences between our approach and related zero-shot semantic segmentation methods based on CLIP.}\label{tab: different}
\vspace{-2mm}
\begin{centering}
\scalebox{0.82}{
\begin{tabular}{c|cccc}
\hline 
\multirow{2}{*}{Methods} & Need an extra & CLIP as an image & Can do\tabularnewline
 & image encoder? & -level classifier? & inductive?\tabularnewline
\hline 
zsseg\cite{xu2021simple} & \Checkmark{} & \Checkmark{} & \Checkmark{}\tabularnewline
ZegFormer\cite{ding2022decoupling} & \Checkmark{} & \Checkmark{} & \Checkmark{}\tabularnewline
MaskCLIP+\cite{zhou2022maskclip} & \Checkmark{} & \XSolidBrush{} & \XSolidBrush{}\tabularnewline
\hline 
\textbf{ZegCLIP (Ours)} & \XSolidBrush{} & \XSolidBrush{} & \Checkmark{}\tabularnewline
\hline 
\end{tabular}
}
\par\end{centering}
\vspace{-3mm}
\end{table}

\noindent\textbf{Design 1:} Using \textbf{Deep Prompt Tuning (DPT)} instead of fine-tuning or fixing for the CLIP image encode. We find that fine-tuning could lead to overfitting to seen classes while prompt tuning prefers to retain the inherent zero-shot capacity of CLIP.

\noindent \textbf{Design 2:} Applying \textbf{Non-mutually Exclusive Loss (NEL)} function when performing pixel-level classification but generating the posterior probability of one class independent of the logits of other classes.

\noindent \textbf{Design 3:} Most importantly and our major innovation --- introducing a \textbf{Relationship Descriptor (RD)} to incorporate the image-level prior into text embedding
before matching text-patch embeddings from CLIP can significantly prevent the model from overfitting to the seen classes.

By incorporating those three designs into our \textbf{one-stage} baseline, we create a simple-but-effective zero-shot semantic segmentation model named \textbf{ZegCLIP}. Tab.~\ref{tab: different} summarizes the differences between our proposed method and existing approaches based on CLIP. More details can be found in Appendix. We conduct extensive experiments on three public datasets and show that our method outperforms the state-of-the-art methods by a large margin in both the ``inductive'' and ``transductive'' settings.

\begin{figure*}[t]
\begin{center}
\includegraphics[width=0.95\linewidth]{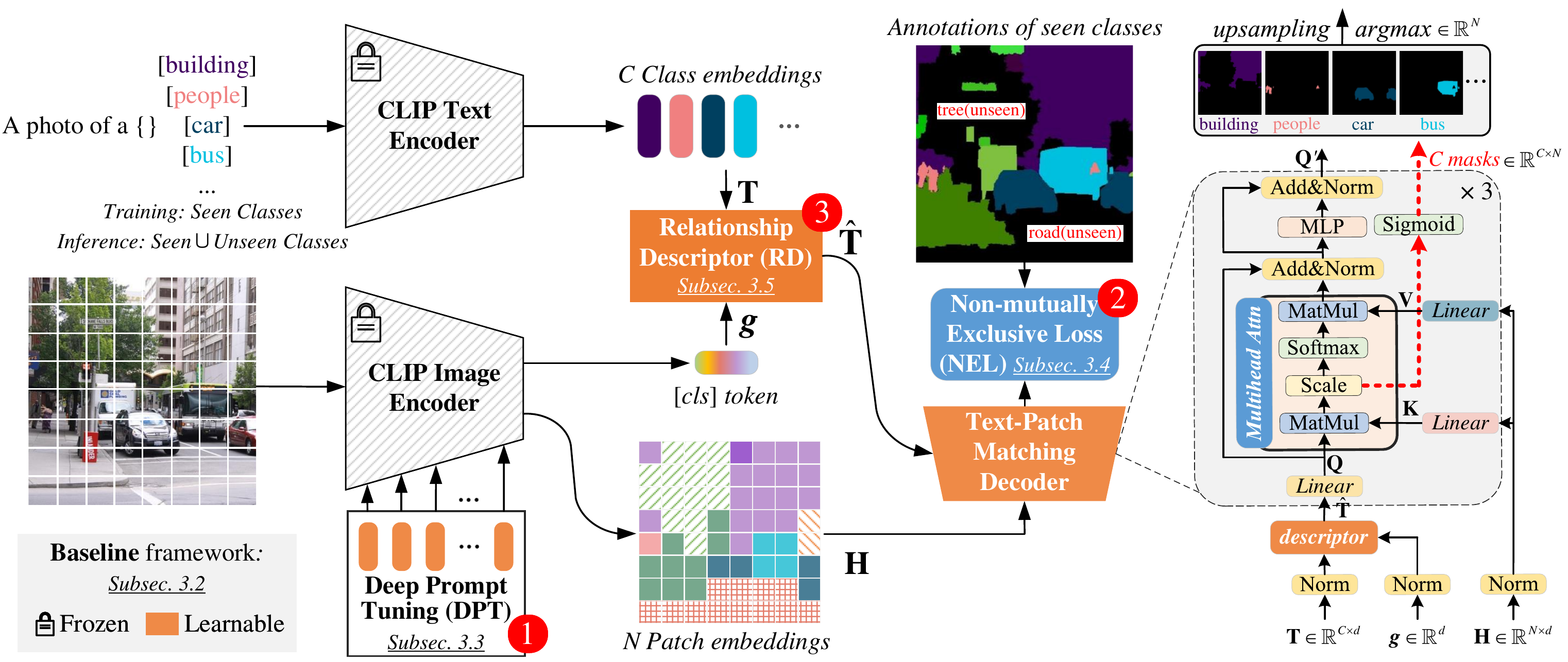}
\end{center}
\vspace{-5mm}
\caption{Overall of our proposed \textbf{ZegCLIP}. Our method modifies a \textbf{One-Stage Baseline} framework of matching text and patch embeddings from CLIP to generate semantic masks. The key contribution of our work is three simple-but-effective designs (labeled as the red circles 1,2,3 in the figure). Incorporating these three designs into our proposed one-stage baseline framework can upgrade the poorly performed baseline method to a strong zero-shot segmentation model.}
\vspace{-3mm}
\label{fig:overall}
\end{figure*}

\section{Related Works}\label{sec:related work}
\textbf{Pre-trained Vision Language Models} \cite{su2019vl,li2019visualbert,radford2021learning,jia2021scaling} which connects the image representation with text embedding have achieved significant performance on various downstream tasks, like image retrieval \cite{liu2021image}, dense prediction \cite{rao2022denseclip}, visual referring expression \cite{wang2022cris}, visual question answering \cite{jiang2022finetuning} and so on. CLIP \cite{radford2021learning} as one of the most popular vision-language models is pre-trained via contrastive learning on 400 million text-image pairs and shows powerful zero-shot classification ability. In this work, we explore how to adapt CLIP's pre-trained vision-language knowledge from image-level into pixel-level prediction efficiently.

\textbf{Semantic Segmentation} as a fundamental dense prediction task in computer vision requires annotating each pixel of the input image. Previous work follows two principles: 1) Directly considering semantic segmentation as per-pixel classification task  \cite{long2015fully,zhang2018context,strudel2021segmenter,xie2021segformer,zheng2021rethinking}; 2) Decoupling the mask generation and the semantic classification~\cite{cheng2021perpixel,cheng2022masked,zhang2022segvit}. Both methods achieve significant progress with a pre-defined closed set of semantic classes.

\textbf{Zero-shot Semantic Segmentation} remains an important but challenging task due to the inevitable imbalance problem in seen classes. The target model is required to segment unseen classes after training on seen classes with annotated labels. Previous works like SPNet\cite{xian2019semantic}, ZS3\cite{bucher2019zero}, CaGNet\cite{gu2020context}, SIGN\cite{cheng2021sign}, Joint\cite{baek2021exploiting} and STRICT\cite{pastore2021closer} follow the strategy that improves the generalization ability of semantic mapping from seen classes to unseen classes.

As the popular pre-trained vision-language model CLIP shows the powerful ability of zero-shot classification, it has been also applied in zero-shot semantic segmentation recently. Zegformer\cite{ding2022decoupling} and zsseg\cite{xu2021simple} develop an extensive proposal generator and use CLIP to classify each region and then ensemble the predicting results. Although it remains CLIP's zero-shot ability at the image level, the computational cost increases inevitably due to classifying each proposal. MaskCLIP+\cite{zhou2022maskclip} creatively applies CLIP to generate pseudo annotations on novel classes (``transductive'') for self-training. It has achieved competitive performance but will be invalid in ``inductive'' setting where the names of unseen classes in inference are unavailable while training.

\textbf{Fine-tuning and Prompt Tuning} are different methods to update the parameters of pre-trained model when adopt on downstream tasks. Intuitively, fine-tuning \cite{hinterstoisser2018pre,wang2022bridging} is able to achieve significant performance in fully supervised learning tasks but is challenged by transfer learning. Recently, prompting tuning that freezes the pre-trained model while introducing a set of learnable prompts \cite{brown2020language} provides an alternative way. It has achieved satisfied performance in both NLP tasks \cite{li2021prefix,qin2021learning,liu2022p} and CV tasks \cite{yao2021cpt,zhou2022learning,xing2022class}. Visual Prompt Tuning \cite{jia2022visual} proposes an effective solution that inserts trainable parameters in each layer of the transformer.

\textbf{Self-training in Zero-shot Segmentation} introduces another setting of zero-shot semantic segmentation called ``transductive''. Unlike the traditional ``inductive'' setting where the novel class names and annotations are both unavailable in the training stage, \cite{pastore2021closer} proposed that self-training via pseudo labels on unlabeled pixels benefits solving the imbalance problem. In such ``transductive'' situation, both the ground truth of seen classes, as well as pseudo labels of unseen classes, will be utilized to supervise the target model for self-training \cite{gu2020context,xu2021simple,zhou2022maskclip} which can achieve better performance.

\section{Method}
\subsection{Problem Definition} \label{sec: problem}
Our proposed method follows the generalized zero-shot semantic segmentation (GZLSS) \cite{xian2019semantic}, which requires to segment both seen classes $\mathcal{C}^{s}$ and unseen classes $\mathcal{C}^{u}$ after only training on a dataset with pixel-annotations of seen part. In the training stage, the model generates per-pixel classification results from the semantic description of all seen classes. In the testing stage, the model is expected to produce segmentation results for both known and novel classes. Note that $\mathcal{C}^{s}\cap\mathcal{C}^{u}=\oslash$ and the labels of $\mathcal{C}^{u}$ are unavailable while training. The key problem of zero-shot segmentation is that merely training on seen classes inevitably leads to overbias on known categories while inference.

This naturally corresponds to the ``inductive'' zero-shot segmentation setting, in which both unseen class names and images are not accessible during training. Besides ``inductive'' zero-shot segmentation, there is a ``transductive'' zero-shot learning setting, which assumes that the names of unseen classes are known before the testing stage. They \cite{gu2020context,zhou2022maskclip} suppose that the training images include the unseen objects, and only ground truth masks for these regions are not available. Our method can easily be extended to both settings and achieve excellent performance.

\subsection{Baseline: One-stage Text-Patch Matching}
As the large-scale pre-trained model CLIP shows impressive zero-shot classification ability, recent methods explore applying CLIP to zero-shot segmentation by proposing a \textbf{two-stage} paradigm. 
In stage 1, they train a class-agnostic generator and then leverage CLIP as a zero-shot image-level classifier by matching the similarity between text embeddings and [cls] token of each proposal in stage 2. While effective, such a design requires two image encoding processes and brings expensive computational overhead.

To simplify the two-stage pipeline when adapting CLIP to zero-shot semantic segmentation, in this work, we aim to cope with the critical problem that how to transfer CLIP's powerful generalization capability from image to pixel level classification effectively. Motivated by the observation of recent work \cite{zhou2022maskclip} that the text embedding can be implicitly matchable to patch-level image embeddings, we build a \textbf{one-stage} baseline 
by adding a vanilla light-weight transformer as a decoder inspired by \cite{dosovitskiy2020image,zhang2022segvit}. Then we formulate semantic segmentation as a matching problem between a representative class query and the image patch features.

Formally, let's denote the $C$
class embeddings
as $\mathbf{T}=[\mathbf{t}^{1},\mathbf{t}^{2},...,\mathbf{t}^{C}]\in\mathbb{R}^{C\times d}$, with $d$ is the feature dimension of CLIP model, $\mathbf{t}^{i}$ representing the $i$th class, and the $N$ patch tokens of an image as $\mathbf{H}=[\mathbf{h}_{1},\mathbf{h}_{2},...,\mathbf{h}_{N}]\in\mathbb{R}^{N\times d}$, with $\mathbf{h}_{j}$ denoting the $j$th patch.

Then we apply linear projections $\phi$ to generate $\mathbf{Q}$($query$), $\mathbf{K}$($key$) and $\mathbf{V}$($value$) as:
\begin{equation}
\begin{array}{c}
\mathbf{Q}=\phi_{q}(\mathbf{T})\in\mathbb{R}^{C\times d}\\
\mathbf{K}=\phi_{k}(\mathbf{H})\in\mathbb{R}^{N\times d},\mathbf{V}=\phi_{v}(\mathbf{H})\in\mathbb{R}^{N\times d}.
\end{array}
\end{equation}
The semantic masks could be calculated by the scaled dot-product attention which is the intermediate product of the multi-head attention model (MHA):
\vspace{-1mm}
\begin{equation}
\label{eq:decoder}
\mathtt{Masks}=\frac{\mathbf{Q}\mathbf{K}^{T}}{\sqrt{d_{k}}}\in\mathbb{R}^{C\times N},
\vspace{-1mm}
\end{equation}
where $\sqrt{d_{k}}$ is the dimension of the keys as a scaling factor and the final segmentation results are obtained by applying $\mathtt{Argmax}$ operation on the class dimension of $\mathtt{Masks}$. 
The detailed architecture of the decoder has been shown on the right of Fig.~\ref{fig:overall}, which consists of three layers of transformers.

\noindent \textbf{How to update the CLIP image encoder:} The patch feature representation is generated by the CLIP image encode.
How to update the CLIP image encoder, e.g., how to calculate $\mathbf{H}$, is an important factor. In our baseline approaches, we consider $\mathbf{H}$ is obtained from a parameter fixed CLIP or a parameter tunable CLIP, denoted as Baseline-Fix and Baseline-FT separately. Later in SubSec.~\ref{sec:vpt}, we will discuss Deep Prompt Tuning (DPT), which turns out to be a better way to adapt the CLIP for zero-shot segmentation. 

\noindent \textbf{How to train the segmentation model:} To properly train the decoder and (optionally) the CLIP model, we apply the commonly used softmax operator to convert the logits calculate from Eq.~\ref{eq:decoder} to the posterior probability. Exclusive Loss (EL) like Cross-entropy is then used as the objective function. Later in SubSec.~\ref{subsec: loss}, we will point out this seemly straightforward strategy can be potentially harmful for generalization. 

\noindent \textbf{Design of query embedding $\mathbf{T}$: } The query embeddings $\mathbf{T}$ are the key to our approach. In our baseline model, we use the embeddings from the CLIP text encoder. However, as will be pointed out in SubSec.~\ref{subsec: adaption}, such a choice might cause severe overfitting and we propose a relationship descriptor by using the relationship between text and image token as class queries. In SubSec.~\ref{subsec: ablation}-A, we further explore other choices of $\mathbf{T}$. Please refer to the relevant sections for more details.

As shown in Fig.~\ref{fig:effectiveness}-(1)(2) and Tab.~\ref{tab: comparison of baseline and designs}, our baseline model turns out to perform poorly in practice, especially for the unseen classes. It seems that the model overfits severely to the seen classes and forgets the zero-shot learning capability of CLIP during training. Fortunately, we identify\(\) three simple-but-effective designs that can dramatically alleviate this issue, as will be described in the subsequent sections.

\subsection{Design 1: Deep Prompt Tuning (DPT)}
\label{sec:vpt}

\begin{figure}[t]
\begin{center}
\includegraphics[width=1.0\linewidth]{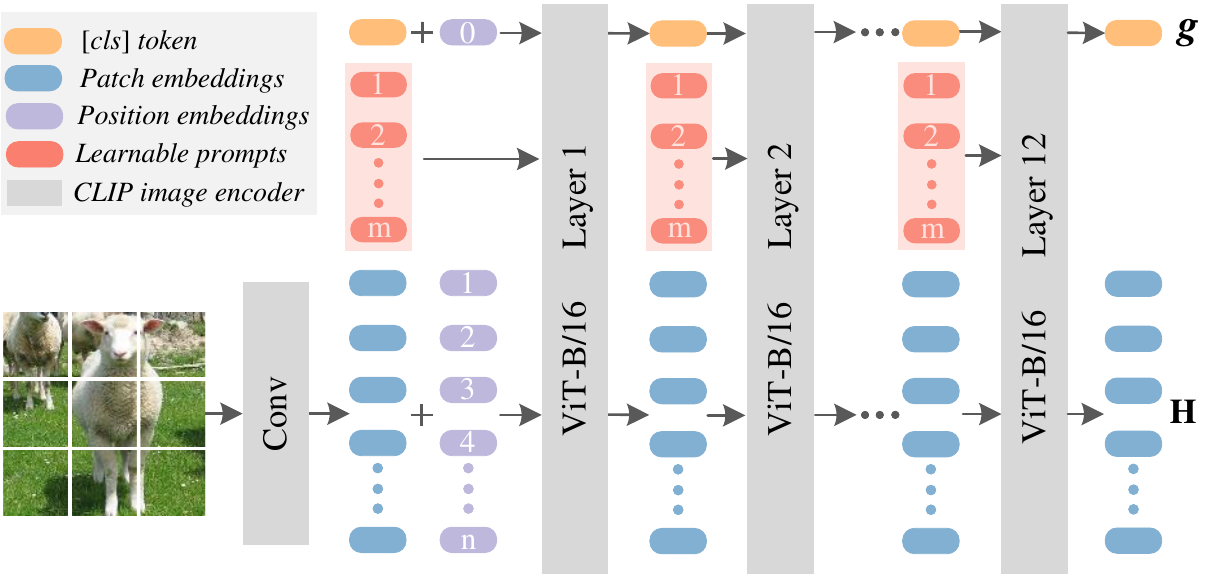}
\end{center}
\vspace{-5mm}
\caption{The architecture of deep prompt tuning.}
\label{fig:prompt}
\vspace{-5mm}
\end{figure}

The first design modification is to use deep prompt tuning (DPT) \cite{zhou2022learning} for the CLIP backbone rather than fine-tuning CLIP. As described in Sec.~\ref{sec:related work}, prompt tuning \cite{xing2022class} is a recently proposed scheme for adapting a pre-trained transformer model to a target domain. It has become a competitive alternative to fine-tuning in the transfer learning setting \cite{jia2022visual}. In this work, we explore deep prompt tuning to adapt CLIP for zero-shot image segmentation. Prompt tuning fixes the original parameters of CLIP and adds learnable prompt tokens as additional input for each layer. Since the zero-shot segmentation model is trained on a limited number of seen classes, directly fine-tuning the model tends to overfit the seen classes as the model parameters are adjusted to optimize the loss only for the seen classes. Consequently, knowledge learned for vision concepts unseen from the training set might be discarded in fine-tuning. Prompt tuning could potentially alleviate this issue since the original parameters are intact during training.

Formally, we denote the input embeddings from the $l$-th MHA module of the ViT-based image encoder in CLIP as $\{\mathbf{g}^l,\mathbf{h}^l_1,\mathbf{h}^l_2,\cdots,\mathbf{h}^l_N\}$, where $\mathbf{g}^l$ denotes the [CLS] token embedding and $\mathbf{H}^{l}=\{{\mathbf{h}^l_1,\mathbf{h}^l_2,\cdots,\mathbf{h}^l_N\}}$ denotes the image patch embeddings. Deep prompt tuning appends learnable tokens $\mathbf{P}^l=\{\mathbf{p}^l_1,\mathbf{p}^l_2,\cdots,\mathbf{p}^l_M \}$ to the above token sequence in each ViT layer of CLIP image encoder. Then the $l$-th MHA module process the input token as:
\vspace{-1mm}
\begin{equation}
[\mathbf{g}^{l},\,\_\,,\mathbf{H}^{l}]=\mathtt{Layer}^{l}([\mathbf{g}^{l-1},\mathbf{P}^{l-1},\mathbf{H}^{l-1}])
\vspace{-1mm}
\end{equation}
where the output embeddings of $\{\mathbf{p}^l_1,\cdots,\mathbf{p}^l_M \}$ are discarded (denoted as $\_$) and will not feed into the next layer. Therefore, $\{\mathbf{p}^l_1,\mathbf{p}^l_2,\cdots,\mathbf{p}^l_M\}$ merely acts as a set of learnable parameters to adapt the MHA model. 

As shown in Fig.~\ref{fig:effectiveness}-(3) and Tab.~\ref{tab: comparison of baseline and designs}, compared with fine-tuning, deep prompt tuning on CLIP image encoder 
can achieve similar performance on seen classes but improve the segmentation results on unseen significantly.

\subsection{Design 2: Non-mutually Exclusive Loss (NEL)} \label{subsec: loss}

The general practice of semantic segmentation models treats it as a per-pixel multi-way classification problem, and Softmax operation is used to calculate the posterior probability, followed by using mutually Exclusive Loss (\textbf{EL}) like Cross Entropy as the loss function. However, 
Softmax essentially assumes a mutually exclusive relationship between the to-be-classified classes: a pixel has to belong to one of the classes of interest. Thus only the relative strength of logits, i.e., the ratio of logits, matters for posterior probability calculation. However, when applying the model to unseen classes, the class space will be different from the training scenario, making the logit of an unseen class poorly calibrated with the other unseen classes.

To handle this issue, we suggest avoiding the mutual-exclusive mechanism during training time and using Non-mutually Exclusive Loss (\textbf{NEL}), more specifically, Sigmoid and Binary Cross Entropy (BCE) loss to ensure the segmentation result for each class is independently generated. In addition, we use the focal loss \cite{lin2017focal} variation of the BCE loss and combine it with an extra dice loss \cite{milletari2016v,zhao2020rethinking} as previous work \cite{cheng2021perpixel}:
\vspace{-2mm}
\begin{equation}
\mathcal{L}_{\mathtt{focal}}=-\frac{1}{\mathrm{hw}}\sum_{i=1}^{\mathrm{hw}}(1-y_{i})^{\gamma}\times\hat{y}\mathrm{log}(y_{i})+y_{i}^{\gamma}\times(1-\hat{y_{i}})\mathrm{log}(1-y_{i}),
\vspace{-1mm}
\end{equation}
\begin{equation}
\mathcal{L}_{\mathtt{dice}}=1-\frac{2\sum_{i=1}^{\mathrm{hw}}y_{i}\hat{y}_{i}}{\sum_{i=1}^{\mathrm{hw}}y_{i}^{2}+\sum_{i=1}^{\mathrm{hw}}\hat{y}_{i}^{2}},
\end{equation}
\begin{equation}
\label{eq: total loss}
\mathcal{L}=\alpha\cdot\mathcal{L}_{\mathtt{focal}}+\beta\cdot\mathcal{L}_{\mathtt{dice}},
\vspace{-1mm}
\end{equation}
where $\gamma=2$ balances hard and easy samples and $\{\alpha,\beta\}$ are coefficients to combine focal loss and dice loss.
As shown in Fig.~\ref{fig:effectiveness}-(4) and Tab. \ref{tab: comparison of baseline and designs}, compared with CE, BCE performs better on unseen classes, and the focal loss and dice loss on BCE \cite{cheng2021perpixel} boost the performance furthermore as demonstrated in SubSec.~\ref{subsec: ablation}-C.

\subsection{Design 3: Relationship Descriptor (RD)}\label{subsec: adaption}

In the above design, the class embeddings extracted from the CLIP text encoder will match against the patch embeddings from the CLIP image encoder in the decoder head. While being quite intuitive, we find this design could lead to severe overfitting. We postulate that this is because the matching capability between the text query and image patterns is only trained on the seen-class datasets. 

This motivates us to incorporate the matching capability learned from the original CLIP training into the transformer decoder. Specifically, we noticed that CLIP calculates the matching score between a text prompt and an image by

\vspace{-2mm}
\begin{equation}
{\mathbf{t}^c}^\top \mathbf{g} = \sum_j^d t^c_j g_j = \sum_j^d r^c_j,
\vspace{-2mm}
\end{equation}
where $\mathbf{t}^c \in \mathbb{R}^d$ denotes a text embedding for the $c$th class and $t_j$ is its $j$-th dimension; $\mathbf{g}$ is the image embedding ([cls] token); $r^c_j = t^c_j g_j$. We postulate that $\mathbf{r}^c = [r^c_1, r^c_2,\cdots,r^c_d]$ characterizes how the image and text, i.e., the text prompt for representing class $c$, are matched, and call it text-image \textbf{Relationship Descriptor (RD)} denoted as $\mathbf{R}\in\mathbb{R}^{C\times d}$ for all $C$ classes in this work. 
Then we concatenate RDs with the original text embedding $\mathbf{T}\in\mathbb{R}^{C\times d}$ as image-specific text queries $\hat{\mathbf{T}}=\{\mathbf{\hat{t}}^1,\mathbf{\hat{t}}^2,...,\mathbf{\hat{t}}^C\}\in\mathbb{R}^{C\times 2d}$  for the transformer decoder. Specifically, with this scheme, the input text query of transformer decoder for each class becomes:
\vspace{-2mm}
\begin{equation}
\mathbf{\hat{t}}=concat[\mathbf{r},\mathbf{t}]=concat[\mathbf{t}\odot\mathbf{g},\mathbf{t}],
\vspace{-2mm}
\end{equation}where $\odot$ is the Hadamard product. Note that, we apply learnable linear projection layers on both $\hat{\mathbf{T}}$ and $\mathbf{H}$ to make them has the same feature dimension.

We compare the effectiveness of applying the relationship description shown in Fig.~\ref{fig:effectiveness}-(5) as well as in Tab.~\ref{tab: comparison of baseline and designs}. As seen, it can dramatically improve the segmentation results on both seen and unseen categories. In SubSec.~\ref{subsec: ablation}-A, we further discuss the effect of using different combination formats of element-wise operations for text-image matching in the relationship descriptor and text queries $\mathbf{\hat{T}}$.

\section{Experiments}

\begin{table*}[h]
\renewcommand\arraystretch{0.95}
\caption{Comparison with the state-of-the-art methods on PASCAL VOC 2012, COCO-Stuff 164K, and PASCAL Context datasets. ``ST'' represents applying self-training via generating pseudo labels on all unlabeled pixels, while ``\Cross''+``ST'' denotes that pseudo labels are merely annotated on unseen pixels excluding the ignore part.}
\vspace{-3mm}
\label{tab: results}
\centering{}%
\resizebox{\textwidth}{38mm}{
\begin{tabular}{c|cccc|cccc|cccc}

\hline 
\multicolumn{1}{c|}{\multirow{2}{*}{Methods}} & \multicolumn{4}{c|}{\textbf{PASCAL VOC 2012}} & \multicolumn{4}{c|}{\textbf{COCO-Stuff 164K}} & \multicolumn{4}{c}{\textbf{PASCAL Context}}\tabularnewline
\cline{2-13} \cline{3-13} \cline{4-13} \cline{5-13} \cline{6-13} \cline{7-13} \cline{8-13} \cline{9-13} \cline{10-13} \cline{11-13} \cline{12-13} \cline{13-13} 
\multicolumn{1}{c|}{} & \textbf{pAcc} & \textbf{mIoU(S)} & \textbf{mIoU(U)} & \textbf{hIoU} & \textbf{pAcc} & \textbf{mIoU(S)} & \textbf{mIoU(U)} & \textbf{hIoU} & \textbf{pAcc} & \textbf{mIoU(S)} & \textbf{mIoU(U)} & \textbf{hIoU}\tabularnewline
\hline 
\multicolumn{13}{l}{\textbf{\textit{Inductive}}}\tabularnewline
\hline 
\multicolumn{1}{c|}{SPNet\cite{xian2019semantic}} & - & 78.0 & 15.6 & 26.1 & - & 35.2 & 8.7 & 14.0 & - & - & - & -\tabularnewline
\multicolumn{1}{c|}{ZS3\cite{bucher2019zero}} & - & 77.3 & 17.7 & 28.7 & - & 34.7 & 9.5 & 15.0 & 52.8 & 20.8 & 12.7 & 15.8\tabularnewline
\multicolumn{1}{c|}{CaGNet\cite{gu2020context}} & 80.7 & 78.4 & 26.6 & 39.7 & 56.6 & 33.5 & 12.2 & 18.2 & - & 24.1 & 18.5 & 21.2\tabularnewline
\multicolumn{1}{c|}{SIGN\cite{cheng2021sign}} & - & 75.4 & 28.9 & 41.7 & - & 32.3 & 15.5 & 20.9 & - & - & - & -\tabularnewline
\multicolumn{1}{c|}{Joint\cite{baek2021exploiting}} & - & 77.7 & 32.5 & 45.9 & - & - & - & - & - & 33.0 & 14.9 & 20.5\tabularnewline
\multicolumn{1}{c|}{ZegFormer\cite{ding2022decoupling}} & - & 86.4 & 63.6 & 73.3 & - & 36.6 & 33.2 & 34.8 & - & - & - & -\tabularnewline
\multicolumn{1}{c|}{zsseg\cite{xu2021simple}} & 90.0 & 83.5 & 72.5 & 77.5 & 60.3 & 39.3 & 36.3 & 37.8 & - & - & - & -\tabularnewline
\multicolumn{1}{c|}{\textbf{ZegCLIP (Ours)}} & \textbf{94.6} & \textbf{91.9} & \textbf{77.8} & \textbf{84.3} & \textbf{62.0} & \textbf{40.2} & \textbf{41.4} & \textbf{40.8} & \textbf{76.2} & \textbf{46.0} & \textbf{54.6} & \textbf{49.9}\tabularnewline
\hline 
\multicolumn{13}{l}{\textbf{\textit{Transductive}}}\tabularnewline
\hline 
SPNet+ST\cite{xian2019semantic} & - & 77.8 & 25.8 & 38.8 & - & 34.6 & 26.9 & 30.3 & - & - & - & -\tabularnewline
ZS5 \cite{bucher2019zero} & - & 78.0 & 21.2 & 33.3 & - & 34.9 & 10.6 & 16.2  & 49.5 & 27.0 & 20.7 & 23.4\tabularnewline
CaGNet+ST\cite{gu2020context} & 81.6 & 78.6 & 30.3 & 43.7 & 56.8 & 35.6 & 13.4 & 19.5 & - & - & - & -\tabularnewline
STRICT\cite{pastore2021closer}& - & 82.7 & 35.6 & 49.8 & - & 35.3 & 30.3 & 34.8 & - & - & - & -\tabularnewline
zsseg+ST\cite{xu2021simple} & 88.7 & 79.2 & 78.1 & 79.3 & 63.8 & 39.6 & 43.6 & 41.5 & - & - & - & -\tabularnewline
\textbf{ZegCLIP+ST (Ours)} & \textbf{95.1} & \textbf{91.8} & \textbf{82.2} & \textbf{86.7} & \textbf{68.8} & \textbf{40.6} & \textbf{54.8} & \textbf{46.6} & \textbf{77.2} & \textbf{46.6} & \textbf{65.4} & \textbf{54.4}\tabularnewline
\hline 
\Cross MaskCLIP+\cite{zhou2022maskclip} & - & 88.8 & 86.1 & 87.4 & - & 38.1 & 54.7 & 45.0 & - & 44.4 & 66.7 & 53.3\tabularnewline
\textbf{\Cross ZegCLIP+ST (Ours)} & \textbf{96.2} & \textbf{92.3} & \textbf{89.9} & \textbf{91.1} & \textbf{69.2} & \textbf{40.7} & \textbf{59.9} & \textbf{48.5} & \textbf{77.3} & \textbf{46.8} & \textbf{68.5} & \textbf{55.6}\tabularnewline
\hline 
\multicolumn{13}{l}{\textbf{\textit{Fully Supervised}}}\tabularnewline
\hline 
\textbf{ZegCLIP (Ours)} & \textbf{96.3} & \textbf{92.4} & \textbf{90.9} & \textbf{91.6} & \textbf{69.9} & \textbf{40.7} & \textbf{63.2} & \textbf{49.6} & \textbf{77.5} & \textbf{46.5} & \textbf{78.7} & \textbf{56.9}\tabularnewline
\hline 

\end{tabular}
}
\vspace{-3mm}
\end{table*}

\subsection{Datasets} 
To evaluate the effectiveness of our proposed method, we conducted extensive experiments on three public benchmark datasets, including PASCAL VOC 2012, COCO-Stuff 164K, and PASCAL Context. The unseen classes of each dataset are shown in the Appendix according to previous works \cite{gu2020context,xu2021simple,ding2022decoupling,zhou2022maskclip}. The details of the datasets are elaborated as follows:

\noindent \textbf{PASCAL VOC 2012} contains 10,582 augmented images for training and 1,449 for validation. We also ignore the ``background'' category and use 15 classes as the seen part and 5 classes as the unseen part.

\noindent \textbf{COCO-Stuff 164K} is a large-scale dataset that contains 171 categories with 118,287 images for training and 5,000 for testing. The whole dataset is divided into 156 seen classes and 15 unseen classes. 

\noindent \textbf{PASCAL Context} includes 60 classes with 4,996 for training and 5,104 for testing. The dataset is divided into 50 known classes (including ``background'') and the rest 10 classes as used as unseen classes in the test set.

\subsection{Evaluation Metrics}

Following previous works, we measure pixel-wise classification accuracy (pAcc) and the mean of class-wise intersection over union (mIoU) on both seen and unseen classes, denoted as $mIoU(S)$ and $mIoU(U)$, respectively. We also evaluate the harmonic mean IoU ($hIoU$) among seen and unseen classes.

\subsection{Implementation Details}
\vspace{-2mm}
Our proposed method is implemented based on the open-source toolbox MMSegmentation\cite{mmseg2020} with PyTorch 1.10.1. All experiments we provided are based on the pre-trained CLIP ViT-B/16 model and conducted on 4 Tesla V100 GPUs, and the batch size is set to 16 with 512x512 as the resolution of images. For ``inductive'' zero-shot learning, the total training iterations are 20K for PASCAL VOC 2012, 40K for PASCAL Context, and 80K for COCO-Stuff 164K. In the ``transductive'' setting, we train our ZegCLIP model on seen classes in the first half of training iterations and then apply self-training via generating pseudo labels in the rest of iterations. The optimizer is set to AdamW with the default training schedule in the MMSeg toolbox.

\subsection{Comparison with State-of-the-art methods}
\vspace{-2mm}
To demonstrate the effectiveness of our method, the evaluation results compared with previous state-of-the-art methods are reported in Tab.~\ref{tab: results}. We also provide the fully supervised learning results as the upper bound to show the performance gap between fully-supervised segmentation and zero-shot segmentation results on unseen classes. The qualitative results on COCO-Stuff 164K are shown in Fig.~\ref{fig:visualization} and more visualization results are provided in Appendix.
 
From Tab.~\ref{tab: results}, we can see that our proposed method achieves significant performance under the ``inductive'' setting and outperforms previous works, especially for unseen classes. This clearly demonstrates the superior generalization capability of our method over the previous approaches. In addition, we also find our approach excels in the ``transductive'' setting, although our method is not specifically designed for that setting. As seen, our model dramatically improves the performance on unseen classes while maintaining excellent performance on seen part after self-training. 

Fig.~\ref{fig:visualization} shows the segmentation results of the Baseline-Fix version and our proposed ZegCLIP on seen and unseen classes. After applying our designs, ZegCLIP shows impressive segmentation ability on both seen and unseen classes and can clearly distinguish similar unseen categories, for example, the unseen ``tree'', ``grass'' and ``playingfield'' categories in (1).

In addition, to demonstrate the efficiency of our proposed method, we compare the number of learnable parameters and inference speed between our one-stage ZegCLIP and typical two-stage method Zegformer\cite{ding2022decoupling} in Tab.~\ref{tab: efficiency}. 
Our proposed method achieves significant performance while only requiring approximately 14M learnable parameters with 117 Flops(G) on average which is only \textbf{23\%} and \textbf{6\%} of \cite{ding2022decoupling}, respectively. In terms of frames Per Second (FPS), our method can achieve a speedup of about \textbf{5 times} faster during inference compared with two-stage method.

\begin{table}[t]
\caption{{\small{}Efficiency comparison with different metrics.  All models are evaluated on a single 1080Ti GPU. \#Params represents the number of learnable parameters in the whole framework.}}
\label{tab: efficiency}
\vspace{-3mm}
\centering{}%

\resizebox{7.5cm}{!}{
\begin{tabular}{cc|ccc}
\hline 
Datasets & Methods & \#Params(M) $\downarrow$ & Flops(G) $\downarrow$ & FPS $\uparrow$\tabularnewline
\hline 
\multirow{2}{*}{VOC} & ZegFormer\cite{ding2022decoupling} & 60.3 & 1829.3 & 1.7\tabularnewline
 & \textbf{ZegCLIP} & \textbf{13.8} & \textbf{110.4} & \textbf{9.0}\tabularnewline
\hline 
\multirow{2}{*}{COCO} & ZegFormer\cite{ding2022decoupling} & 60.3 & 1875.1 & 1.5\tabularnewline
 & \textbf{ZegCLIP} & \textbf{14.6} & \textbf{123.9} & \textbf{6.7}\tabularnewline
\hline 
\end{tabular}
}\vspace{-6mm}
\end{table}

\begin{figure*}[t]
\begin{center}
\includegraphics[width=0.95\linewidth]{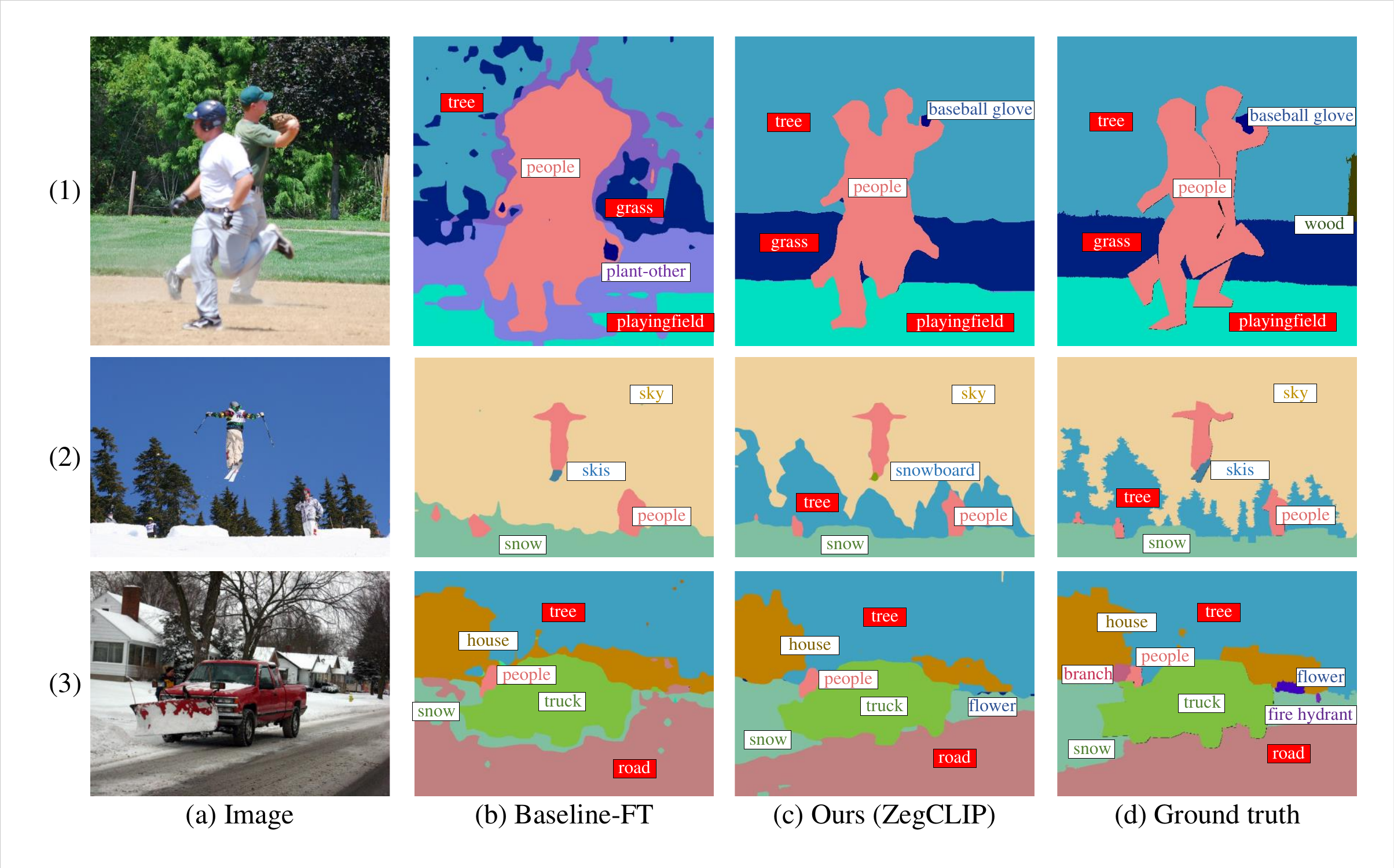}
\end{center}
\vspace{-5mm}
\caption{Qualitative results on COCO-Stuff 164K. (a) are the original testing images; (b) represent the performance of our proposed one-stage baseline (fine-tuning the image encoder); (c) are the visualization results of our proposed ZegCLIP; (d) are the ground truths of each image. Note that the white and \colorbox {red}{\textcolor{white}{red}} tags represent seen and unseen classes separately.}
\vspace{-5mm}
\label{fig:visualization}
\end{figure*}

\subsection{Ablation Study}\label{subsec: ablation}
\vspace{-2mm}

\noindent \textbf{A. Effect of different formats of text query $\mathbf{\hat{t}}$} \label{subsubsec: q}

In this work, we propose an important design called Relationship Descriptor (RD) as described in SubSec.~\ref{subsec: adaption}. In this module, we combine the queried text embeddings $\mathbf{T}$ with image [cls] token $\mathbf{g}$ extract from CLIP to generate text-image description $\mathbf{\hat{T}}$ before feeding the queries into the segment decoder. Such image-specific queries can dramatically improve the zero-shot segmentation performance on unseen classes. To further explore the effect of different element-wise operations in the relationship descriptor, we conduct various formats of  $\mathbf{\hat{T}}$ on VOC and report the results in Tab.~\ref{tab: q formats}. For the image-specific text query of each class $\mathbf{\hat{t}}$, it can be formulated as shown in the second column. 

We can see, the dot product and the absolute difference between text embedding $\mathbf{t}$ and [cls] token $\mathbf{g}$ can provide more general information, but the sum and concatenate operations perform poorly on both seen and unseen classes. This is understandable since dot product and absolute difference characterize the relationship between the image and text encoding, which is in line with the interpretation made in SubSec.~\ref{subsec: adaption}. The \textcolor{red}{red line} in Tab.~\ref{tab: q formats} is the format of $\mathbf{\hat{t}}$ we finally chose in our ZegCLIP model.

\begin{table}[h]
\renewcommand\arraystretch{0.9}
\caption{{\small{}Effect of different formats of text queries $\mathbf{\hat{t}}$.}}\label{tab: q formats}
\vspace{-3mm}
\scalebox{0.85}{
\centering{}%
\begin{tabular}{c|c|cccc}
\hline 
\textbf{dim} & \textbf{format of $\mathbf{\hat{t}}$} & \textbf{pAcc} & \textbf{mIoU(S)} & \textbf{mIoU(U)} & \textbf{hIoU}\tabularnewline
\hline 
\multirow{5}{*}{512} & \textbf{t} & 86.8 & 89.5 & 33.7 & 49.0\tabularnewline
 & \textbf{t}$\odot$\textbf{g} & 93.1 & 90.2 & 68.4 & 77.8\tabularnewline
 & |\textbf{t}-\textbf{g}| & 92.4 & 90.6 & 64.2 & 75.1\tabularnewline
 & \textbf{t}-\textbf{g} & 88.7 & 87.9 & 46.5 & 60.8\tabularnewline
 & \textbf{t}+\textbf{g} & 82.2 & 89.9 & 13.9 & 24.1\tabularnewline
\hline 
\multirow{6}{*}{512{*}2} & {[}\textbf{t}, \textbf{g}{]} & 88.9 & 88.8 & 39.3 & 54.5\tabularnewline
 & \textbf{\textcolor{red}{{[}t$\odot$g, t{]}}} & \textbf{\textcolor{red}{94.6}} & \textbf{\textcolor{red}{91.9}} & \textbf{\textcolor{red}{77.8}} & \textbf{\textcolor{red}{84.3}}\tabularnewline
 & {[}|\textbf{t}-\textbf{g}|, \textbf{t}{]} & 90.9 & 91.5 & 54.2 & 68.1\tabularnewline
 & {[}\textbf{t$\odot$g}, \textbf{t}+\textbf{g}{]} & 88.3 & 90.0 & 38.0 & 53.4\tabularnewline
 & {[}\textbf{t}+\textbf{g}, \textbf{t}{]} & 82.8 & 89.4 & 20.7 & 33.6\tabularnewline
 & {[}\textbf{t}$\odot$\textbf{g}, |\textbf{t}-\textbf{g}|{]} & 94.1 & 91.2 & 73.9 & 81.6\tabularnewline
\hline 
512{*}3 & {[}\textbf{t}$\odot$\textbf{g}, |\textbf{t}-\textbf{g}|,\textbf{ t}{]} & 93.4 & 91.6 & 67.3 & 77.6\tabularnewline
\hline 
\end{tabular}
}
\vspace{-5mm}
\end{table}

\noindent \textbf{B. Detailed results of applying designs on baseline}

\begin{table*}[t]
\renewcommand\arraystretch{0.85}
\caption{{\small{}Quantitative results on VOC and COCO dataset to demonstrate the effectiveness of our proposed three designs.}}\label{tab: comparison of baseline and designs}\label{tab: designs}
\vspace{-3mm}
\centering{}%
\scalebox{1.0}{
\begin{tabular}{c|cccc|cccc}
\hline 
\multirow{2}{*}{\textbf{\small{}method}} & \multicolumn{4}{c|}{\textbf{\small{}PASCAL VOC 2012}} & \multicolumn{4}{c}{\textbf{\small{}COCO-Stuff 164K}}\tabularnewline
\cline{2-9} \cline{3-9} \cline{4-9} \cline{5-9} \cline{6-9} \cline{7-9} \cline{8-9} \cline{9-9} 
 & \textbf{\small{}pAcc} & \textbf{\small{}mIoU(S)} & \textbf{\small{}mIoU(U)} & \textbf{\small{}hIoU} & \textbf{\small{}pAcc} & \textbf{\small{}mIoU(S)} & \textbf{\small{}mIoU(U)} & \textbf{\small{}hIoU}\tabularnewline
\hline 
{\small{}Baseline-Fix} & {\small{}69.3} & {\small{}71.1} & {\small{}16.3} & {\small{}26.5} & {\small{}33.3} & {\small{}17.1} & {\small{}15.4} & {\small{}16.2}\tabularnewline
{\small{}Baseline-Fix + NEL} & {\small{}85.5} & {\small{}85.2} & {\small{}36.6} & {\small{}51.2} & {\small{}52.4} & {\small{}31.7} & {\small{}20.8} & {\small{}25.1}\tabularnewline
{\small{}Baseline-Fix + RD} & {\small{}86.0} & {\small{}82.5} & {\small{}46.6} & {\small{}59.6} & {\small{}41.0} & {\small{}23.3} & {\small{}23.4} & {\small{}23.3}\tabularnewline
{\small{}Baseline-Fix + NEL + RD} & {\small{}89.6} & {\small{}83.3} & {\small{}66.4} & {\small{}73.9} & {\small{}53.7} & {\small{}32.3} & {\small{}32.5} & {\small{}32.4}\tabularnewline
\hline 
{\small{}Baseline-FT} & {\small{}77.3} & {\small{}76.5} & {\small{}13.8} & {\small{}23.4} & {\small{}48.4} & {\small{}32.4} & {\small{}17.5} & {\small{}22.7}\tabularnewline
{\small{}Baseline-FT + NEL} & {\small{}83.8} & {\small{}84.1} & {\small{}27.5} & {\small{}41.4} & {\small{}56.5} & {\small{}39.9} & {\small{}25.4} & {\small{}31.0}\tabularnewline
{\small{}Baseline-FT + RD} & {\small{}79.4} & {\small{}77.8} & {\small{}20.7} & {\small{}32.7} & {\small{}54.0} & {\small{}39.6} & {\small{}22.4} & {\small{}28.6}\tabularnewline
{\small{}Baseline-FT + NEL + RD} & {\small{}89.6} & {\small{}90.2} & {\small{}42.4} & {\small{}57.7} & {\small{}60.2} & \textbf{\small{}42.7} & {\small{}22.3} & {\small{}29.3}\tabularnewline
\hline 
{\small{}Baseline-DPT} & {\small{}76.2} & {\small{}75.9} & {\small{}28.3} & {\small{}41.2} & {\small{}39.0} & {\small{}22.5} & {\small{}17.5} & {\small{}19.7}\tabularnewline
{\small{}Baseline-DPT + NEL} & {\small{}89.2} & {\small{}89.9} & {\small{}40.4} & {\small{}55.7} & {\small{}58.5} & {\small{}38.0} & {\small{}27.4} & {\small{}31.8}\tabularnewline
{\small{}Baseline-DPT + RD} & {\small{}85.5} & {\small{}81.0} & {\small{}55.2} & {\small{}65.7} & {\small{}46.4} & {\small{}28.4} & {\small{}27.8} & {\small{}28.1}\tabularnewline
\textbf{\small{}Baseline-DPT + NEL + RD (ZegCLIP)} & \textbf{\small{}94.6} & \textbf{\small{}91.9} & \textbf{\small{}77.8} & \textbf{\small{}84.3} & \textbf{\small{}62.0} & {\small{}40.2} & \textbf{\small{}41.4} & \textbf{\small{}40.8}\tabularnewline
\hline 
\end{tabular}}
\end{table*}

\begin{figure*}[t]
\begin{center}
\vspace{-3.5mm}
\includegraphics[width=0.95\linewidth]{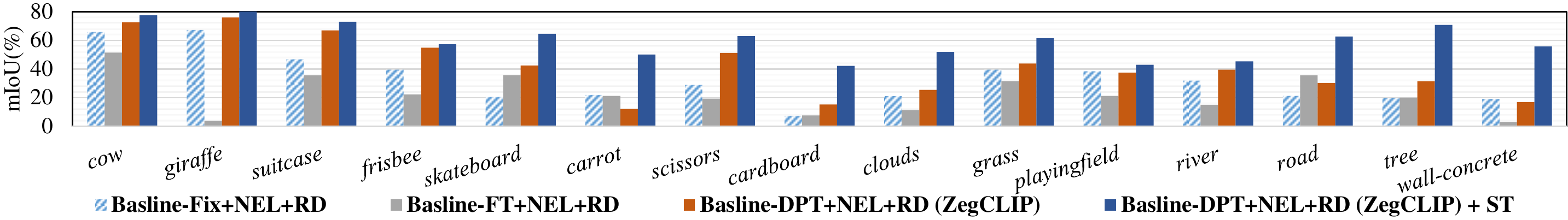}
\end{center}
\vspace{-7.5mm}
\caption{Detailed performance on unseen classes of COCO datasets. Note that ``ST'' represents self-training  in ``transductive'' setting.}
\label{fig:cooc-unsee}
\vspace{-3mm}
\end{figure*}

To demonstrate the effectiveness of our proposed designs, we further report the improvements of applying designs on our Baseline model in Tab.~\ref{tab: comparison of baseline and designs}. It shows that fixed CLIP (Baseline-Fix) with a learnable segment decoder fails to achieve satisfactory performance on both seen and unseen classes due to weak image representation for the dense prediction task. Meanwhile, fine-tuning CLIP (Baseline-FT) performs better on seen classes but deteriorates on unseen classes dramatically. It seems that the zero-shot transfer learning ability of CLIP is destroyed after updating on seen classes. Changing objective functions from mutual exclusive into \textbf{Non-mutually Exclusive Loss (NEL, Design 2)} also leads to significant performance boost both in seen and unseen scenarios. The most dramatic improvement comes from \textbf{Relationship Descriptor (RD, Design 3)} which almost doubles the unseen performance in many cases. Finally, we noticed that \textbf{Deep Prompt Tuning (DPT, Design 1)} works well with NEL (Design 2) and RD (Design 3). If we replace DPT by fine-tuning CLIP and combine it with NEL and RD, the issue of overfitting seen classes remain. In conclusion, our proposed method can ensure semantic segmentation performance while maintaining the powerful ability of CLIP and achieving excellent performance on seen and unseen classes simultaneously.

\vspace{1mm}
\noindent \textbf{C. Effect of advanced loss function}

As we described in Subsec \ref{subsec: loss}, we propose that applying non-mutually exclusive objective functions like using binary cross entropy (BCE) with sigmoid performs better on zero-shot semantic segmentation. To better handle the imbalance problem among categories, we further combine BCE with focal loss\cite{lin2017focal} and dice loss\cite{li2019dice} according to \cite{cheng2021perpixel,zhang2022segvit} as Eq.~\ref{eq: total loss}, and the results of applying such two versions of the loss function, denoted as  ``plain'' and ``plus'', are reported separately in Tab.~\ref{tab: plus loss}. We can see that the ``plus'' loss function achieves better performance on both seen and unseen classes on VOC, COCO, and Context datasets.

\vspace{-2mm}
\begin{table}[h]
\caption{Comparison of introducing advanced loss function. Note that ``plain'' represents merely Binary Cross Entropy (BCE), while ``plus'' means adding focal loss on BCE and dice loss}\label{tab: plus loss}
\renewcommand\arraystretch{0.82}
\vspace{-3mm}
\begin{centering}
{\small{}}%
\begin{tabular}{c|c|cccc}
\hline 
\textbf{\small{}dataset} & \textbf{\small{}loss} & \textbf{\small{}pAcc} & \textbf{\small{}mIoU(S)} & \textbf{\small{}mIoU(U)} & \textbf{\small{}hIoU}\tabularnewline
\hline 
\multirow{2}{*}{\textbf{\small{}VOC}} & {\small{}plain} & {\small{}93.4} & {\small{}89.7} & {\small{}73.6} & {\small{}80.9}\tabularnewline
 & \textbf{\small{}plus} & \textbf{\small{}94.6} & \textbf{\small{}91.9} & \textbf{\small{}77.8} & \textbf{\small{}84.3}\tabularnewline
\hline 
\multirow{2}{*}{\textbf{\small{}COCO}} & {\small{}plain} & {\small{}59.8} & {\small{}38.8} & {\small{}39.0} & {\small{}38.9}\tabularnewline
 & \textbf{\small{}plus} & \textbf{\small{}62.0} & \textbf{\small{}40.2} & \textbf{\small{}41.4} & \textbf{\small{}40.8}\tabularnewline
\hline 
\multirow{2}{*}{\textbf{\small{}Context}} & {\small{}plain} & {\small{}75.3} & {\small{}43.5} & {\small{}50.0} & {\small{}46.5}\tabularnewline
 & \textbf{\small{}plus} & \textbf{\small{}76.2} & \textbf{\small{}46.0} & \textbf{\small{}54.6} & \textbf{\small{}49.9}\tabularnewline
\hline 
\end{tabular}{\small\par}
\vspace{-3mm}
\par\end{centering}
\end{table}

\vspace{1mm}
\noindent \textbf{D. Generalization ability to other datasets}

To further explore the generalization ability of our proposed method, we conduct extra experiments in Tab.~\ref{tab: generalization}. We apply the pre-trained model of the source dataset via supervised learning on seen classes and evaluate the segmentation results on both seen and unseen classes of target datasets. Our method shows better cross-domain generalization capability compared with the latest related work Zegformer which is also based on the CLIP model.

\vspace{-2mm}
\begin{table}[h]
\caption{Generalization ability to other datasets.} \label{tab: generalization}
\vspace{-3mm}
\centering{}%
\scalebox{0.85}{
\begin{tabular}{c|c|c|ccc}
\hline 
\textbf{\small{}source} & \textbf{\small{}target} & \textbf{\small{}method} & \textbf{\small{}pAcc} & \textbf{\small{}mIoU} & \textbf{\small{}mAcc}\tabularnewline
\hline 
\multirow{6}{*}{\textbf{\small{}COCO}} & \multirow{3}{*}{\textbf{\small{}Context}} & {\small{}Zegformer\cite{ding2022decoupling}} & {\small{}56.8} & {\small{}36.1} & {\small{}64.0}\tabularnewline
 &  & {\small{}ZegCLIP} & {\small{}60.9} & {\small{}41.2} & {\small{}68.4}\tabularnewline
 &  & \textbf{\small{}\Cross ZegCLIP+ST} & \textbf{\small{}68.4} & \textbf{\small{}45.8} & \textbf{\small{}70.9}\tabularnewline
\cline{2-6} \cline{3-6} \cline{4-6} \cline{5-6} \cline{6-6} 
 & \multirow{3}{*}{\textbf{\small{}VOC}} & {\small{}Zegformer\cite{ding2022decoupling}} & {\small{}92.8} & {\small{}85.6} & {\small{}92.7}\tabularnewline
 &  & {\small{}ZegCLIP} & {\small{}96.9} & {\small{}93.6} & {\small{}96.4}\tabularnewline
 &  & \textbf{\small{}\Cross ZegCLIP+ST} & \textbf{\small{}97.2} & \textbf{\small{}94.1} & \textbf{\small{}96.7}\tabularnewline
\hline 
\end{tabular}
}
\vspace{-3mm}
\end{table}

\vspace{-2mm}
\section{Conclusion}
\vspace{-3mm}
In this work, we propose an efficient one-stage straightforward zero-shot semantic segmentation method based on the pre-trained vision-language CLIP. To transfer the image-wise classification ability to dense prediction tasks while maintaining the advanced zero-shot knowledge, we figure out three designs to achieve competitive results on seen classes while extremely improving the performance on novel classes. Our proposed method relay on text embeddings as queries that are very flexible to cope with both ``inductive'' and ``transductive'' zero-shot settings. To demonstrate the effectiveness of our method, we conduct extensive performance on three public benchmark datasets and outperform previous state-of-the-art methods. Meanwhile, our one-stage framework shows about 5 times faster compared with two-stage methods while inference. 
In general, our work explores how to leverage the pre-trained vision-language model CLIP into semantic segmentation and successfully utilize its zero-shot knowledge in downstream tasks which may provide inspiration for future research.

\noindent \textbf{Acknowledgement.} This work was done in Adelaide Intelligence Research (AIR) Lab and Lingqiao Liu is supported by the Centre of Augmented Reasoning (CAR).

\balance
{\small
\bibliographystyle{ieee_fullname}
\bibliography{egbib}
}
\clearpage

\appendix

\section*{Appendix}

\section{Effect of the number of deep prompt tokens}

\begin{figure}[h]
\begin{center}
\includegraphics[width=1.0\linewidth]{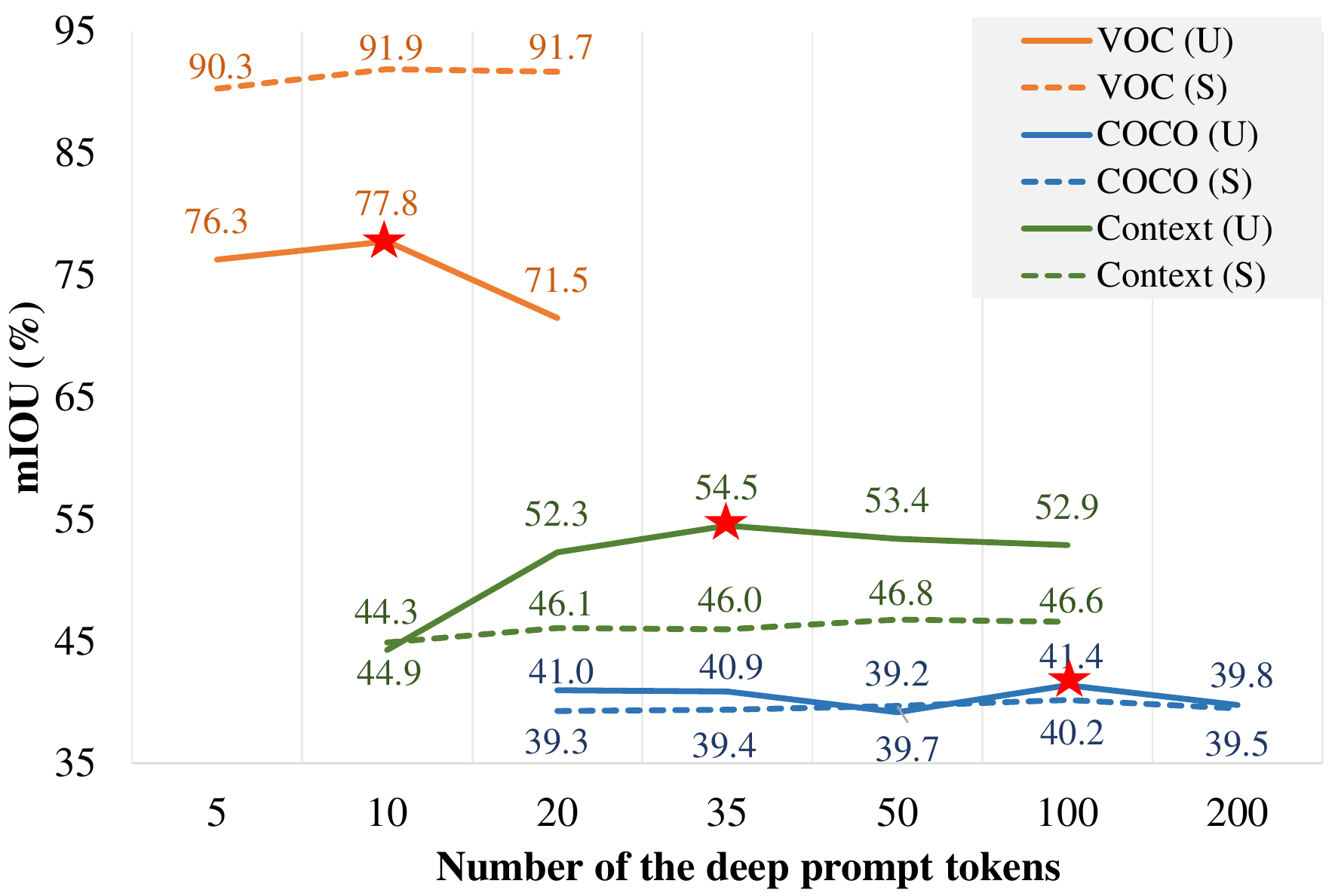}

\end{center}
\vspace{-5mm}
\caption{The quantitative results of applying the different numbers of deep prompt tokens. Note that ``S'' and ``U'' represent seen and unseen classes separately.}
\label{fig: number of token}
\vspace{-5mm}
\end{figure}

We figure out that the best number of deep prompt tokens varies for different datasets and the detailed results are shown in Fig. \ref{fig: number of token}. For the PASCAL VOC 2012 dataset (VOC) which contains fewer training samples and categories, 10 tokens are enough to obtain significant performance on both seen and unseen classes. However, for large-scale datasets, more deep prompts, i.e., 100 for COCO-Stuff 164K (COCO) and 35 for PASCAL Context (Context), are beneficial to achieve better segmentation performance. In general, the best number of deep prompt tokens increases with the scale of the dataset and the complexity of the per-pixel classification task increases. Meanwhile, using too many visual prompts may be detrimental to our model instead.

\section{Effect of the depth of deep prompt tokens}

\vspace{-3mm}
\begin{table}[h]
\caption{{\small{}Effect of the depth of deep prompt tuning on VOC.}}\label{tab: depth}
\centering{}%
\begin{tabular}{c|cccc}
\hline 
\textbf{layer} & \textbf{pAcc} & \textbf{mIoU(S)} & \textbf{mIoU(U)} & \textbf{hIoU}\tabularnewline
\hline 
1 & 91.4 & 87.5 & 67.8 & 76.4\tabularnewline
1$\rightarrow$3 & 91.7 & 86.7 & 70.2 & 77.6\tabularnewline
1$\rightarrow$6 & 92.7 & 87.8 & 75.3 & 81.1\tabularnewline
1$\rightarrow$9 & 93.3 & 88.9 & 72.4 & 79.8\tabularnewline
\textbf{1$\rightarrow$12} & \textbf{94.6} & \textbf{91.9} & \textbf{77.8} & \textbf{84.3}\tabularnewline
\hline 
10$\rightarrow$12 & 92.5 & 88.3 & 70.9 & 78.6\tabularnewline
7$\rightarrow$12 & 92.5 & 89.0 & 68.0 & 77.1\tabularnewline
4$\rightarrow$12 & 93.6 & 91.5 & 66.9 & 77.3\tabularnewline
\hline 
\end{tabular}
\vspace{-5mm}
\end{table}

Except that the number of deep prompt tokens can impact the performance of zero-shot semantic segmentation, we also conduct extensive experiments on PASCAL VOC 2012 (VOC) to explore the effect of inserting the learnable prompts in different layers of the CLIP image encoder. The quantitative results are reported in Tab. \ref{tab: depth}. For a better explanation, we number the total 12 vision transformer layers in the CLIP image encoder from 1 (``bottom") to 12 (``top") and the layers of inserting prompts are as denoted in the first column of Tab. \ref{tab: depth}. We figure out that adding prompt tokens on ``bottom" layers generally tends to perform better than on ``top" layers. Meanwhile, inserting learnable prompt tokens in each ViT layer (layer=1$\rightarrow$12) achieves the best performance which is also the default setting in our experiments.

\section{Effect of single and multiple text templates}
Following the training details of CLIP, we apply a single template ``\textit{A photo of a} \{\}" on PASCAL VOC 2012 (VOC) and multiple templates on large-scale datasets, i.e.,  COCO-Stuff 164K (COCO) and PASCAL Context (Context), when obtaining the class embeddings from CLIP text encoder. We provide the quantitative results of using single and multiple templates in Tab. \ref{tab: templates} where we can see that multiple descriptions achieve reasonable improvements on both two datasets.

\begin{table}[h]
\caption{Comparison of using single and multiple templates on COCO-Stuff 164K and PASCAL Context datasets.}\label{tab: templates}
\vspace{-3mm}
\begin{centering}
\scalebox{1.0}{
\begin{tabular}{c|c|cccc}
\hline 
\textbf{\small{}dataset} & \textbf{\small{}template} & \textbf{\small{}pAcc} & \textbf{\small{}mIoU(S)} & \textbf{\small{}mIoU(U)} & \textbf{\small{}hIoU}\tabularnewline
\hline 
\multirow{2}{*}{\textbf{\small{}COCO}} & {\small{}single} & {\small{}61.4} & {\small{}39.5} & {\small{}40.6} & {\small{}40.0}\tabularnewline  & \textbf{\small{}multiple} & \textbf{\small{}62.0} & \textbf{\small{}40.2} & \textbf{\small{}41.4} & \textbf{\small{}40.8}\tabularnewline
\hline 
\multirow{2}{*}{\textbf{\small{}Context}} & {\small{}single} & {\small{}75.8} & {\small{}45.1} & {\small{}52.1} & {\small{}48.3}\tabularnewline
  & \textbf{\small{}multiple} & \textbf{\small{}76.2} & \textbf{\small{}46.0} & \textbf{\small{}54.6} & \textbf{\small{}49.9}\tabularnewline
 \hline 
 \end{tabular}{\small\par}
 }\par\end{centering}
 \end{table}

The details of the 15 augmented templates we used on COCO-Stuff 164K and PASCAL Context datasets are:

\vspace{3mm}

`A photo of a \{\}.'

`A photo of a small \{\}.'

`A photo of a medium \{\}.'

`A photo of a large \{\}.'

`This is a photo of a \{\}.'

`This is a photo of a small \{\}.'

`This is a photo of a medium \{\}.'

`This is a photo of a large \{\}.'

`A \{\} in the scene.'

`A photo of a \{\} in the scene.'

`There is a \{\} in the scene.'

`There is the \{\} in the scene.'

`This is a \{\} in the scene.'

`This is the \{\} in the scene.'

`This is one \{\} in the scene.'

\begin{figure*}[h]
\begin{center}
\includegraphics[width=1.0\linewidth]{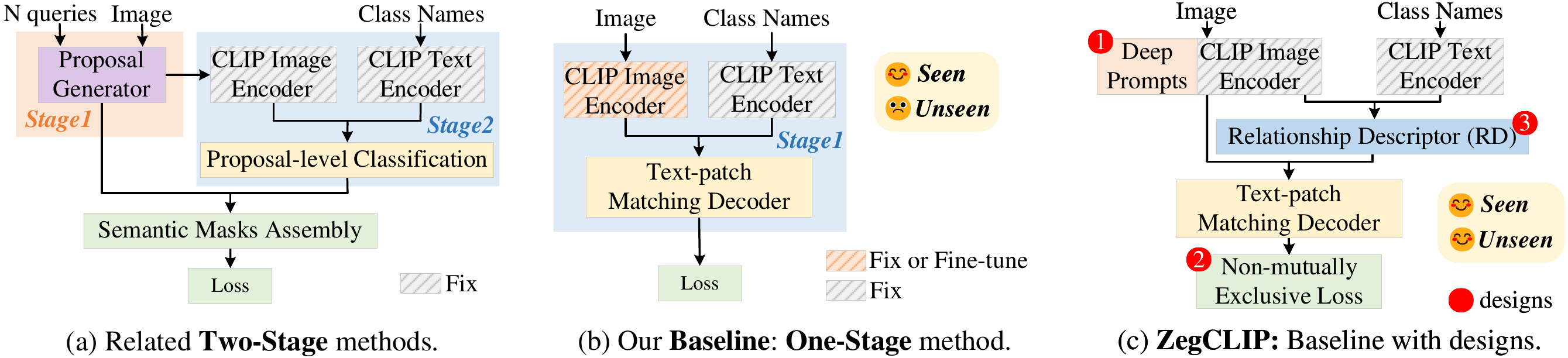}
\end{center}
\vspace{-3mm}
\caption{Brief frameworks of related \textbf{two-stage methods}, our \textbf{one-stage baseline}, and our proposed \textbf{ZegCLIP} model. The happy face in (b)(c) means good performance in seen or unseen classes, while the sad face in (b) represents that the baseline model achieves poor performance in unseen classes.}
\label{fig:two-one-stage}
\end{figure*}

\section{Brief frameworks of related Two-stage and our One-stage method}
As we described above, previous zero-shot semantic segmentation methods based on CLIP follow a \textbf{two-stage} paradigm as shown in Fig. \ref{fig:two-one-stage}-(a). In stage 1, they need to generate abundant class-agnostic region proposals according to the learnable queries. In stage 2, each cropped proposal region will be encoded via CLIP image encoder to utilize its powerful image-level zero-shot classification capability. The CLIP is still used as a zero-shot image-level classifier.

Instead, we propose a simpler-and-efficient \textbf{one-stage} solution as our baseline that directly leverages the feature embedding from CLIP and extends CLIP from image-level classification into pixel-level (or patch-level) as shown in Fig. \ref{fig:two-one-stage}-(b). We introduce a light transformer-based decoder to generate semantic masks by computing the similarities between text-wise and patch-wise embeddings extracted from CLIP. In the baseline, the CLIP text encoder is frozen and the CLIP image encoder can be fixed (Baseline-Fix) or fine-tuned (Baseline-FT).

However, our baseline model still faces the overfitting problem on seen classes. To improve the generalization ability to unseen classes, we propose three important designs on our baseline as shown in \ref{fig:two-one-stage}-(c). After combining three designs, our model ZegCLIP can transfer the zero-shot ability of CLIP from image-level to pixel-level and achieve significant performance on both seen and unseen classes.

In conclusion, in the two-stage methods, $N$ cropped class-agnostic images will be fed into CLIP for image-wise classification which may heavily increase the computational cost. Our proposed one-stage paradigm is simple-but-efficient due to the original image will be encoded only once. The inference speed has been compared in Tab. 3. Our one-stage method ZegCLIP can achieve a speedup of about \textbf{5 times faster} than the two-stage method in the inference stage.

\section{The details of unseen classes names}
For fair comparison, here we provide the detailed unseen class names of PASCAL VOC 2012 (VOC), COCO-Stuff 164K (COCO), and PASCAL Context (Context) dataset in Tab. \ref{tab: names}.

\begin{table}[h]
\caption{The details of unseen class names.}\label{tab: names}
\centering{}{\small{}}%
\begin{tabular}{|c|c|}
\hline 
\textbf{\small{}Dataset} & \textbf{\small{}The name of unseen classes}\tabularnewline
\hline 
{\small{}VOC} & \textit{\small{}pottedplant, sheep, sofa, train, tvmonitor}\tabularnewline
\hline 
\multirow{3}{*}{{\small{}COCO}} & \textit{\small{}cow, giraffe, suitcase, frisbee, skateboard}\tabularnewline
 & \textit{\small{}carrot, scissors, cardboard, clouds, grass}\tabularnewline
 & \textit{\small{}playingfield, river, road, tree, wall concrete}\tabularnewline
\hline 
\multirow{2}{*}{{\small{}Context}} & \textit{\small{}cow, motorbike, sofa, cat, boat, fence}\tabularnewline
 & \textit{\small{}bird, tv monitor, keyboard, aeroplane}\tabularnewline
\hline 
\end{tabular}{\small\par}
\end{table}
\section{More visualization details}

\begin{figure*}[t]
\begin{center}
\includegraphics[width=1.0\linewidth]{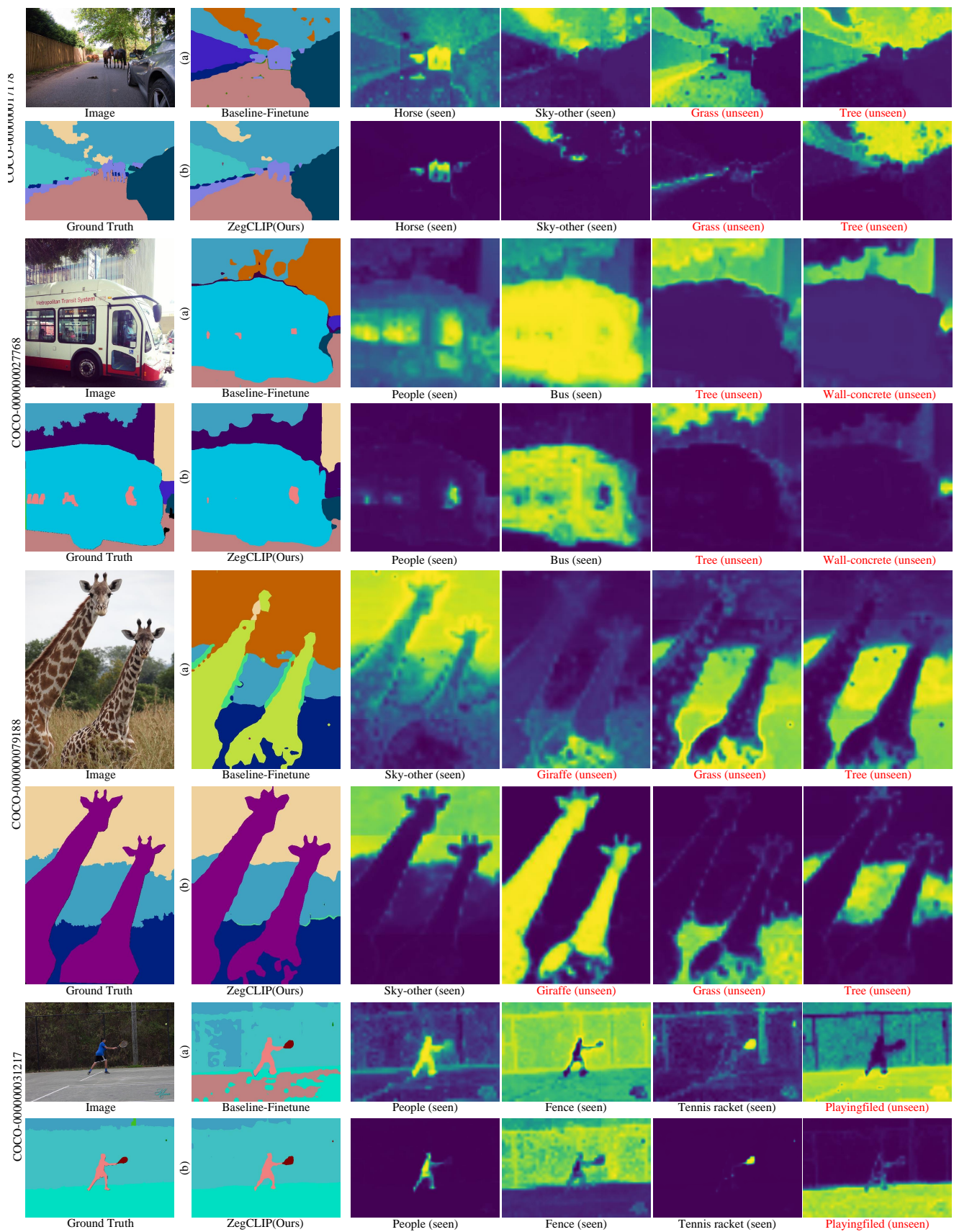}
\end{center}
\vspace{-5mm}
\caption{Visualization of semantic masks of different text query embeddings.} \label{fig: query-vis}
\label{fig:overall}
\end{figure*}

To further demonstrate the effectiveness of our designs on the one-stage baseline (Baseline-FT version), we provide more visualizations including the predicted segmentation results and the semantic masks of different class queries via decoder in Fig. \ref{fig: query-vis}. Note that the class names in red are the novel categories. 

We can see, after applying our proposed designs, the segmentation performance of (b) ZegCLIP has improved on both seen and unseen classes compared with (a) Baseline-FT. Meanwhile, similar unseen classes can be more clearly classified by our ZegCLIP model as shown in the heat maps. For example, in the ``COCO-000000079188'' testing image, although Baseline-FT can classify ``grass'' and ``tree'' (both are unseen classes) correctly in the semantic masks, our ZegCLIP can distinguish these novel classes discriminatively.

\end{document}